\documentclass[10pt,twocolumn,letterpaper]{article}

\usepackage{cvpr}
\usepackage{times}
\usepackage{epsfig}
\usepackage{graphicx}
\usepackage{amsmath}
\usepackage{amssymb}
\usepackage{calc}
\usepackage{dblfloatfix} 
\newlength\myheight
\newlength\mydepth
\settototalheight\myheight{Xygp}
\usepackage[breaklinks=true,bookmarks=false]{hyperref}

\cvprfinalcopy 

\newcommand{\vqa}{{\textsc{VQA }}}
\newcommand*\inlinegraphics[1]{%
  \settototalheight\myheight{Xygp}%
  \settodepth\mydepth{Xygp}%
  \raisebox{-\mydepth}{\includegraphics[height=\myheight]{#1}}%
}

\def\cvprPaperID{4371} 

\begin{document}

\title{Explicit Bias Discovery in Visual Question Answering Models}

\author{Varun Manjunatha, Nirat Saini \& Larry S. Davis \\
Dept. of Computer Science \\
University of Maryland, College Park \\
\texttt{\{varunm@cs, nirat@cs, lsd@umiacs\}.umd.edu}
}

\maketitle

\begin{abstract}
Researchers have observed that Visual Question Answering (VQA) models tend to answer questions by learning statistical biases in the data. For example, their answer to the question ``What is the color of the grass?" is usually ``Green", whereas a question like ``What is the title of the book?" cannot be answered by inferring statistical biases. It is of interest to the community to explicitly discover such biases, both for understanding the behavior of such models, and towards debugging them. Our work address this problem. In a database, we store the words of the question, answer and visual words corresponding to regions of interest in attention maps. By running simple rule mining algorithms on this database, we discover human-interpretable rules which give us unique insight into the behavior of such models. Our results also show examples of unusual behaviors learned by models in attempting VQA tasks.
\end{abstract}

\section{Introduction}
\label{sec:intro}
\begin{figure*}[h]
    \centering
    \includegraphics[width=\textwidth]{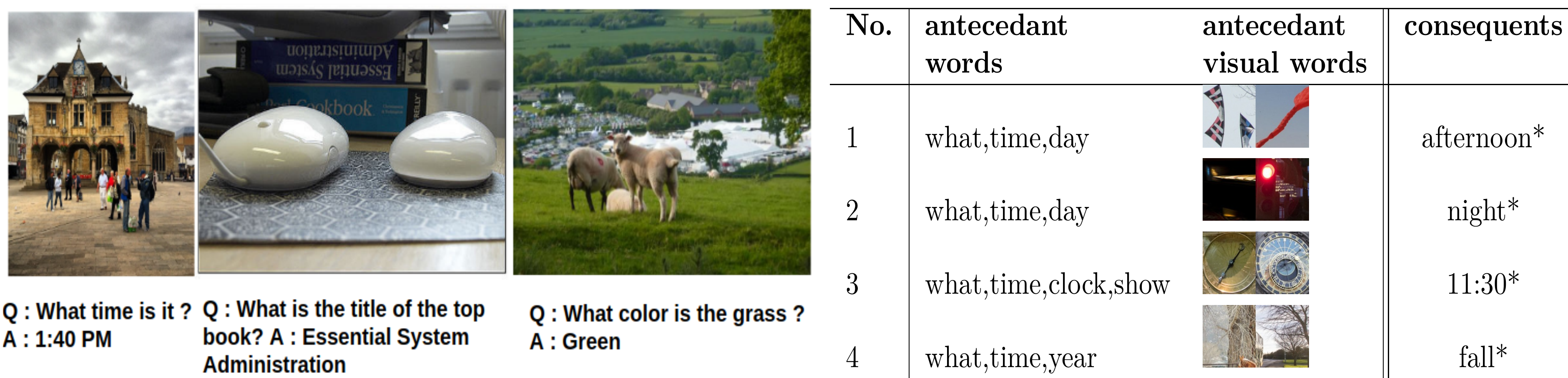}
    \caption{On the left, we show examples of two questions in \vqa which the model requires a ``skill" to answer (such as telling the time, or reading the English language), and a third which can be answered using statistical biases in the data. On the right, we show examples of statistical biases for a set of questions containing the phrase ``What time?" and various visual elements (\emph{antecedents}). Note that each row in this figure represents multiple questions in the \vqa validation set. The * next to the answer (or \emph{consequent}) reminds us that it is from the set of answer words. There are several visual words associated with afternoon and night, but we have provided only two for brevity.}.
    \label{fig:types_of_vqa_and_how_many}
\end{figure*}

In recent years, the problem of Visual Question Answering (\vqa) - the task of answering a question about an image has become a hotbed of research activity in the computer vision community. While there are several publicly available \vqa datasets\cite{antol2015vqa,johnson2016clevr,krishnavisualgenome,MalinowskiF14}, our focus in this paper will be on the dataset provided in \cite{antol2015vqa} and \cite{GoyalKSBP17}, which is the largest natural image-question-answer dataset and the most widely cited. Even so, the narrowed-down version of the \vqa  problem on this dataset is not monolithic - ideally, several different skills are required by a model to answer the various questions. In Figure \ref{fig:types_of_vqa_and_how_many}(left) , a question like ``What time is it?" requires the acquired skill of being able to read the time on  a clock-face, ``What is the title of the top book?" requires an OCR-like ability to read sentences, whereas the question ``What color is the grass?" can be answered largely using statistical biases in the data itself (because frequently in this dataset, grass is green in color). Many models have attempted to solve the problem of \vqa with varying degrees of success, but among them, the vast majority still attempt to solve the \vqa task by exploiting biases in the dataset \cite[etc]{KazemiE17,AndersonTeney,GVQA,MCB,MUTAN}, while a smaller minority address the individual problem types \cite[etc]{NMN,InterpretableCounting, CountingPrith}. 

Keeping the former in mind, in this work, we provide a method to discover and enumerate explicitly, the various biases that are learned by a \vqa model. For example, in Figure \ref{fig:types_of_vqa_and_how_many}(right), we provide examples of some rules learned by a strong baseline \cite{KazemiE17}. The model seems to have learned that if a question contains the words \{What, time, day\} (Eg : ``What time of day is it?") and the accompanying image contains the bright sky (\inlinegraphics{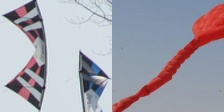}), the model is likely to answer ``afternoon". The model answers ``night" to the same question accompanied with an image containing a ``night-sky" patch (\inlinegraphics{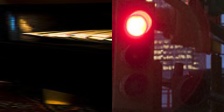}). On the other hand, if it contains a clock face(\inlinegraphics{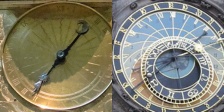}), it tends to answer the question with a time in an ``HH:MM" format, while a question like ``What time of the year?" paired with leafless trees(\inlinegraphics{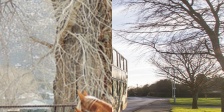}) prompts ``fall" as the answer. The core of our method towards discovering such biases is the classical Apriori algorithm \cite{AgrawalAPriori} which is used to discover rules in large databases - here the \emph{database} refers to the question-words and model responses on the \vqa validation set, which can be mined to produce these rules.

Deep learning algorithms reduce training error by learning biases in the data. This is evident from the observation that validation/test samples from the long tail of a data distribution are hard to solve, simply because similar examples do not occur frequently enough in the training set\cite[etc]{WangNIPS, YangCVPR16}. However, explicitly enumerating these biases in a human-interpretable form is possible only in a handful of problems, such as VQA. \vqa is particularly illustrative because the questions and answers are in human language, while the images (and attention maps) can also be interpreted by humans. \vqa is also interesting because it is a multi-modal problem - both language and vision are required to solve this problem. The language alone (i.e., an image agnostic model) can generate plausible (but often incorrect) answers to \emph{most} questions (as we show in Section \ref{sec:language_only}), but incorporating the image generates more accurate answers. That the language alone is able to produce plausible answers strongly indicates that \vqa models implicitly use simple rules to produce answers - we endeavour in this paper to find an approach that can discover these rules.

Finally, we note that in this work, we do not seek to improve upon the state of the art. We do most of our experiments on the model of \cite{KazemiE17}, which is a strong baseline for this problem. We choose this model because it is simple to train and analyze (Section \ref{sec:baseline}). To concretely summarize, our main contribution is to provide a method that can capture macroscopic rules that a \vqa model ostensibly utilizes to answer questions. To the best of our knowledge, this is the first detailed work that analyzes the \vqa dataset of \cite{GoyalKSBP17} in this manner.

The rest of this paper is arranged as follows : In Section \ref{sec:related}, we discuss related work, specifically those which look into identifying pathological biases in several machine learning problems, and ``debugging" \vqa models. In Section \ref{sec:method}, we discuss details of our method. In Section \ref{sec:experiments}, we provide experimental results and list (in a literal sense) some rules we believe the model is employing to answer questions. We discuss limitations of this method in Section \ref{sec:limitations} and conclude in Section \ref{sec:conclusion}.

\section{Background and Related Work}
\label{sec:related}
The \vqa problem is most often solved as a multi-class classification problem. In this formulation, an image(I) usually fed through a CNN, and a question(Q) fed through a language module like an LSTM \cite{hochreiter1997long} or GRU \cite{GRU}, are jointly mapped to an answer category (``yes", ``no", ``1", ``2", etc). Although the cardinality of the set of all answers given a QI dataset is potentially infinite, researchers have observed that a set of a few thousand (typically 3000 or so) most frequently occurring answers can account for over 90\% of all answers in the \vqa dataset. Further, the evaluation of \vqa in \cite{antol2015vqa} and \cite{GoyalKSBP17} is performed such that an answer receives partial credit if at least one human annotator agreed with the answer, even if it might not be the answer provided by the majority of the annotators. This further encourages the use of a classification based \vqa system that limits the number of answers to the most frequent ones, rather than an answer generation based \vqa system (say, using a decoder LSTM like \cite{vinyals2015show}).

\textbf{On undesirable biases in machine learning models}: Machine learning methods are increasingly being used as tools to calculate credit scores, interest rates, insurance rates, etc, which deeply impact lives of ordinary humans. It is thus vitally important that machine learning models not discriminate on the basis of gender, race, nationality, etc\cite{equality, propublica, DBLP:conf/fat/BuolamwiniG18}. \cite{CisseECCV2018} focus on revealing racial biases in image-based datasets by using adversarial examples. \cite{MenShop} explores data as well as models associated with object classification and visual semantic role labeling for identifying gender biases and their amplification. Further, \cite{Homemaker} shows the presence of gender biases while encoding word embeddings, which is further exacerbated while using those embeddings to make predictions. \cite{HendrixSnowboard} propose an Equalizer model which ensures equal gender probability when making predictions on image captioning tasks.

\textbf{On debugging deep networks}: The seminal work by \cite{Lipton} suggests that the Machine Learning community does not have a good understanding of what it means to interpret a model. In particular, this work expounds \emph{post-hoc interpretability} - interpretation of a model's behavior based on some criteria, such as visualizations of gradients \cite{Gradcam} or attention maps \cite{Xu2015show}, \emph{after} the model has been trained. Locally Interpretable Model Agnostic Explanations (LIME), \cite{lime:kdd16} explain a classifier's behavior at a particular point by perturbing the sample and building a linear model using the perturbations and their predictions. A follow up work \cite{anchors:aaai18} constructs \emph{Anchors}, which are features such that, in an instance where these features hold, a model's prediction does not change. This work is the most similar prior work to ours, and the authors provide a few results on \vqa as well. However, they only assume the existence of a model, and perturb instances of the data, whereas ours assumes the existence of responses to a dataset, but not the model itself. We use standard rule finding algorithms and provide much more detailed results on the \vqa problem. 

\textbf{On debugging \vqa }:\cite{AgrawalBP16} study the behavior of models on the \vqa 1.0 dataset. Through a series of experiments, they show that \vqa models fail on novel instances, tend to answer after only partially reading the question and fail to change their answers across different images. In \cite{GVQA}, recognizing that deep models seem to use a combination of identifying visual concepts and prediction of answers using biases learned from the data, the authors develop a mechanism to disentangle the two. However, they do not explicitly find a way to discover such biases in the first place. In \cite{GoyalKSBP17}, the authors introduce a second, more balanced version of the \vqa dataset that mitigates biases (especially language based ones) in the original dataset. The resulting balanced dataset is christened \vqa 2.0, and is the dataset that our results are reported on. In \cite{kafle2017analysis}, the authors balance yes/no questions (those which indicate the presence or absence of objects), and propose two new evaluation metrics that compensate for forms of dataset bias. 

\section{Method}
\label{sec:method}
\begin{figure*}[h]
    \centering
    \includegraphics[width=\textwidth]{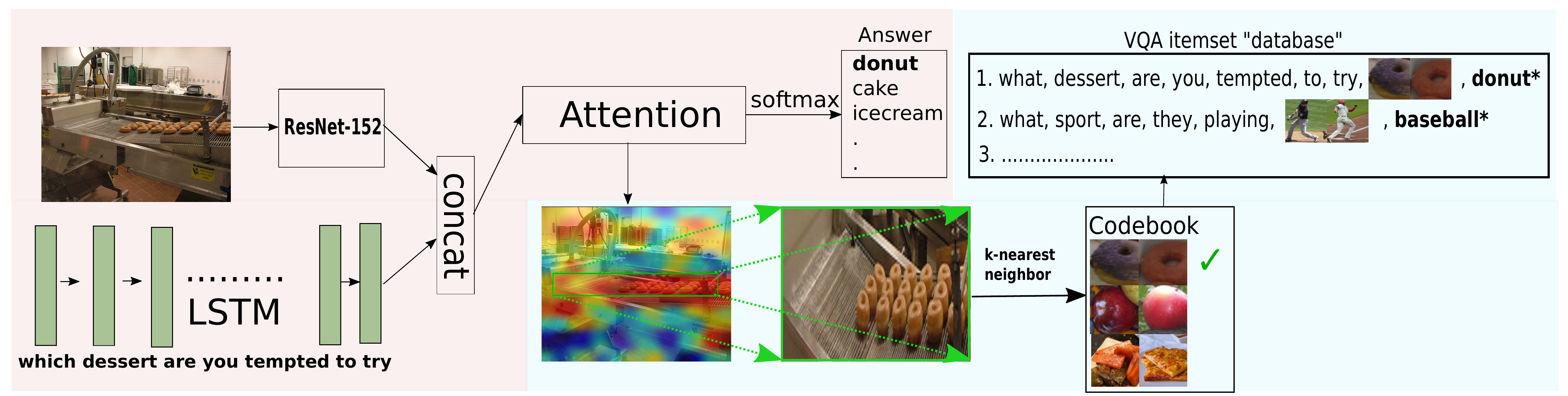}
    \caption{The model from \cite{KazemiE17} tries to answer the question ``Which dessert are you tempted to try?". In doing so, the visual attention focuses on a region of the image which contains donuts. We use the method by \cite{ChenBLL16} to place a bounding box over this region, which maps to a distinct visual word representing \emph{donuts} in our vocabulary. Our database of items thus contains all of the words of the question, the visual word and the answer words. Rules are then extracted using the Apriori algorithm \cite{AgrawalAPriori}}.
    \label{fig:method}
\end{figure*}

We cast our bias discovery task as an instance of the rule mining problem, which we shall describe below. The connection between discovering biases in \vqa and rule mining is as follows : each (Question, Image, Answer) or QI+A triplet can be cast as a transaction in a database, where each word in the question, answer and image patch (or visual word, Section \ref{sec:codebook} and \ref{sec:box}) is akin to an item. There are now three components to our rule mining operation : 
\begin{itemize}
    \item First, a frequent itemset miner picks out a set of all itemsets which occur at least $s$ times in the dataset where $s$ is the support. Because our dataset has over 200,000 questions (the entire \vqa validation set), and the number of items exceeds 40,000 (all question words+all answer words+all visual words), we choose GMiner \cite{ChonH018} due to its speed and efficient GPU implementation. Examples of such frequent itemsets in the context of \vqa include \{what, color, red*\}, \{what, sport, playing\}, where the presence of a * indicates that the word is an answer-word.
    \item Next, a rule miner Apriori \cite{AgrawalAPriori} forms all valid association rules $A \rightarrow C$, such that the rule has a support $>s$ and a confidence $>c$, where the confidence is defined as $\frac{|A \cup C|}{|A|}$. Here, the itemset $A$ is called \emph{antecedent} and the itemset $C$ is called \emph{consequent}. We choose and $c = 0.2$ unless specified otherwise. An example of an association rule is \{what, sport, playing, \inlinegraphics{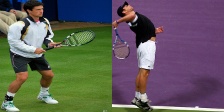}\} $\rightarrow$ \{tennis*\}, which can be interpreted as ``If the question contains the words ---what, sport, playing--- and the accompanying image contains a tennis player, the answer \emph{could} be tennis". 
    \item Finally, a post-processing step removes obviously spurious rules by considering the causal nature of the \vqa problem (i.e., only considering rules that obey : Image/Question $\rightarrow$ Answer). For the purpose of the results in Section \ref{sec:experiments}, we query these rules with search terms like \{What,sport\}.
\end{itemize}
More concretely, let the $i^{th}$ (Image, Question) pair result in the network predicting the answer $a^i$. Let the question itself contain the words $\{w^i_1, w^i_2, ...., w^i_k\}$. Further, while answering the question, let the part of the image that the network shows attention towards correspond to the visual code-word $v^i$ (Section \ref{sec:codebook} and \ref{sec:box}). Then, this QI+A corresponds to the transaction $\{w^i_1, w^i_2, ...., w^k_k, v^i, a^i\}$. By pre-computing and combining question, answer and visual vocabularies, each item in a transaction can be indexed uniquely. This is shown in Figure \ref{fig:method} and explained in greater detail in the following sub-sections. 

\subsection{Baseline Model}
\label{sec:baseline}
The baseline model we use in this work is from \cite{KazemiE17}, which was briefly a state-of-the-art method, yielding higher performance than other, more complicated models. We choose this model for two reasons : first, its simplicity (in other words, an absence of ``bells and whistles") makes it a good test-bed for our method and has been used by other works that explore the behavior of \vqa algorithms \cite{PramodACL, FengRAWR2018}. The second reason is that the performance of this baseline is within 4\% of the state-of-the-art model \cite{AndersonTeney} without using external data or ensembles. We use the implementation of \href{https://github.com/Cyanogenoid/pytorch-vqa}{https://github.com/Cyanogenoid/pytorch-vqa}. A brief description of this model is as follows : The \vqa problem is formulated as a multi-class classification problem (Section \ref{sec:related}). The input to the model is an image and a question, while the output is the answer class with the highest confidence (out of 3000 classes). Resnet-152\cite{Resnet} features are extracted from the image and concatenated with the last hidden state of an LSTM\cite{hochreiter1997long}. The text and visual features are combined to form attention maps which are fed to the softmax (output) layer through two dense layers. In this work, we focus on the second attention map.

\subsection{Visual Codebook Generation}
\label{sec:codebook}
\begin{figure*}[h]
    \centering
    \includegraphics[width=\textwidth]{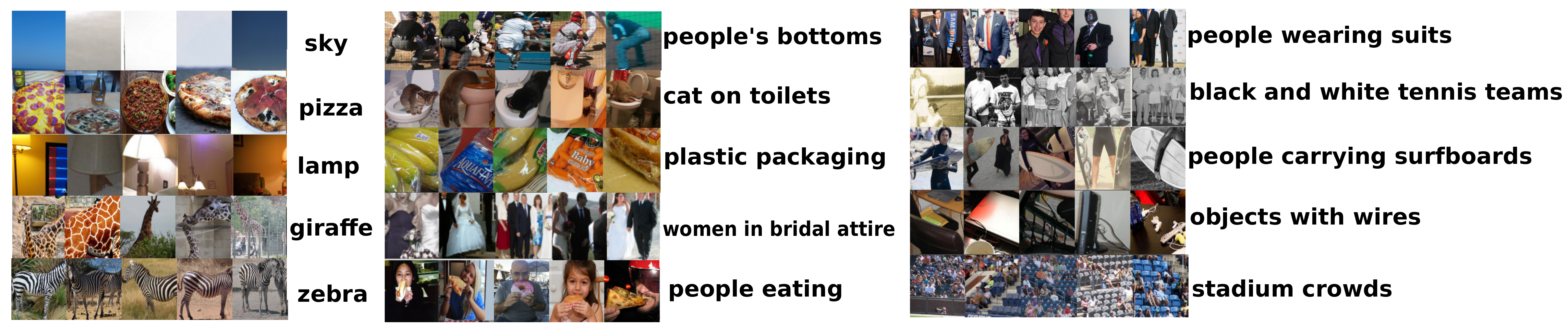}
    \caption{We show visual code-words generated by the method of Section 3.1. In the first (left-most) column, we notice visual code-words corresponding to objects or patches in MSCOCO, but in the latter two columns (on the right) we notice code-words corresponding to more complex visual concepts like ``people eating", ``women in bridal-wear" or ``black-and-white tennis photographs".}
    \label{fig:code-words}
\end{figure*}

We generate the visual codebook using the classical ``feature extraction followed by clustering" technique from \cite{SivicZ03}. First, we use the bounding-box annotations in MSCOCO\cite{mscoco} and COCO-Stuff\cite{caesar2018cvpr} to extract 300,000 patches from the MSCOCO training set. After resizing each of the patches to $224 \times 224$ pixels, we extract ResNet-152\cite{Resnet} features for each of these patches, and cluster them into 1250 clusters using $k$-means clustering\cite{yinyang}. We note in Figure \ref{fig:code-words} that the clusters have both expected and unexpected characteristics beyond ``objectness" and ``stuffness". Expected clusters include dominant objects in the MSCOCO dataset like zebras, giraffes, elephants, cars, buses, trains, people, etc. However, other clusters have textural content, unusual combinations of objects as well as actions. For example, we notice visual words like ``people eating", ``cats standing on toilets", ``people in front of chain link fences", etc, as shown in Figure \ref{fig:code-words}. The presence of these more \emph{eclectic} code-words casts more insight into the model's learning dynamics - we would prefer frequent itemsets containing the visual code-word corresponding to ``people eating" than just ``people" for a QA pair of \emph{(what is she doing?, eating)}.


\subsection{From attention map to bounding box}
\label{sec:box}
\begin{figure*}[h]
    \centering
    \includegraphics[width=\textwidth]{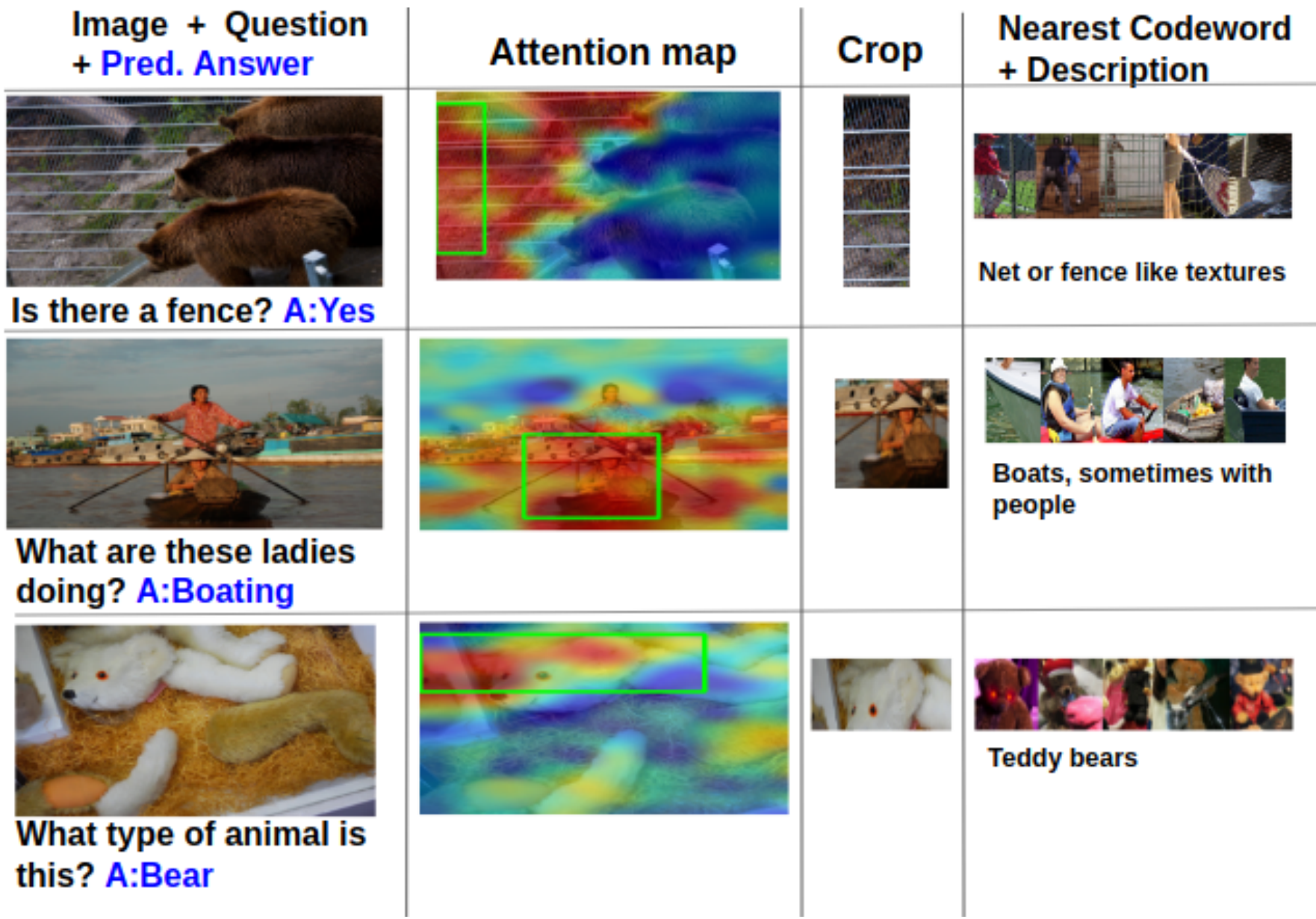}
    \caption{In the first example, critical to answering the question correctly is discovering the presence of a fence (shown in \textcolor{red}{red}) in the attention heat-map. The cropping method of \cite{ChenBLL16} places a conservative box over this region, which corresponds to net-like or fence-like visual code-words like a tennis-net or a baseball batting-cage in the visual codebook. Similarly, in the second example, the attention corresponds to a visual code-word which clearly depicts boats, and in the third example, the attention corresponds to the teddy-bear code-word.}
    \label{fig:croppedcode-words}
\end{figure*}

In this work, we make an assumption that the network focuses on exactly one part of the image, although our method can be easily extended to multiple parts\cite{ChenBLL16}. Following the elucidation of our method in Section \ref{sec:method} and given an attention map, we would like to compute the nearest visual code-word. Doing so requires making the choice of a bounding box that covers enough of the salient parts of the image, cropping and mapping this patch to the visual vocabulary. While there are trainable (deep network based) methods for cropping attention maps \cite{WangDeepCropping}, we instead follow the simpler formulation suggested by \cite{ChenBLL16}, which states that : within an attention-map $G$, given a percentage ratio $\tau$, find the smallest bounding box $B$ which satisfies : 
\[
\sum_{p\epsilon B}{G(p)} \geq \tau \sum_p {G(p)}, \tau \epsilon [0,1]
\]
Since we follow \cite{KazemiE17} who use a ResNet-152 architecture for visual feature extraction, the attention maps are of size $14 \times 14$. It can be shown easily that given a $m \times n$ grid, the number of unique bounding boxes that can be drawn on this grid, i.e., $num\_bboxes$ = $\frac{m \times n \times (m+1) \times (n+1)}{4}$, and when $m=n=14$, $num\_bboxes$ turns out to be 11,025. Because $m(=n)$ is small and fixed in this case, we pre-compute and enumerate all 11,025 bounding boxes and pick the smallest one which encompasses the desired attention, with $\tau=0.3$. The reason behind a conservatively low choice for $\tau$ is that we do not want to crop large regions of the image, which might contain distractor patches. This part of the pipeline is depicted in Figure \ref{fig:croppedcode-words}.

\subsection{Pipeline Summarized}
\label{sec:pipeline}
Now, the pipeline for the experiments (Figure \ref{fig:method}) on the \vqa dataset including images is as follows. We provide as input to the network - an image and a question. We observe the second attention map and use the method of Section \ref{sec:box} to place a tight-fitting bounding-box around those parts of the image that the model attends to. We then extract features on this bounding-box using a ResNet-152 network and perform a $k$-nearest neighbor search (with $k=1$) to obtain its nearest visual word from the vocabulary. The words in the question, visual code-word and predicted answer for the entire validation set are provided as the database of transactions to the frequent itemset miner \cite{ChonH018}, and rules are then obtained using the Apriori algorithm \cite{AgrawalAPriori}.

\section{Experiments}
\label{sec:experiments}
\subsection{Language only statistical biases in \vqa}
\label{sec:language_only}
We show that a large number of statistical biases in VQA are due to language alone. We illustrate this with an obvious example : a language-only model, i.e., one that does not see the image, but still attempts the question, answers about 43\% of the questions correctly on VQA 2.0 validation set and 48\% of the questions correctly on VQA 1.0 validation set\cite{GoyalKSBP17}. However, on a random set of 200 questions from VQA 2.0, we observed empirically that the language-only model answers 88.0\% of questions with a \emph{plausibly correct} answer even with a harsh metric of what \emph{plausible} means. Some of these responses are fairly sophisticated as can be seen in Table \ref{table:language_only_table}. We note, for example, that questions containing ``kind of bird" are met with a species of bird as response, ``What kind of cheese" is answered with a type of cheese, etc. Thus, the model maps out key words or phrases in the question and \emph{ostensibly} tries to map them through a series of rules to answer words. This strongly indicates that these are biases learned from the data, and the ostensible rules can be mined through a rule-mining algorithm.

\begin{table}
\begin{center}
\footnotesize
 \begin{tabular}{|l|p{1.1cm}|p{1.1cm}|} 
 \hline
 Question & Predicted & G.T Ans. \\ [0.5ex] 
 \hline\hline
 What kind of bird is perched on this branch ? & owl & sparrow \\ 
 What does that girl have on her face ? & sunglasses & nothing \\
 What kind of cheese is on pizza ? & mozzarella & mozzarella \\
 What is bench made of ? & wood & wood \\
 \textcolor{red} {What brand of stove is in kitchen ?} & electric & LG \\ 
 \hline
\end{tabular}
\end{center}
\caption{We run a language-only \vqa baseline and note that although only 43\% of the questions are answered correctly in VQA 2.0 (\cite{GoyalKSBP17}), a large number of questions (88\%) in our experiments are answered with plausibly correct responses. For example, ``Sunglasses" would be a perfectly plausible answer to the question ``What does that girl have on her face?" - perhaps even more so than the ground-truth answer (``Nothing"). The \textcolor{red}{last example} shows an implausible answer provided by the model to the question.}
\label{table:language_only_table}
\end{table}

\subsection{Vision+Language statistical biases in \vqa}
\begin{figure*}
    \includegraphics[width=\textwidth,height=.3\textheight]{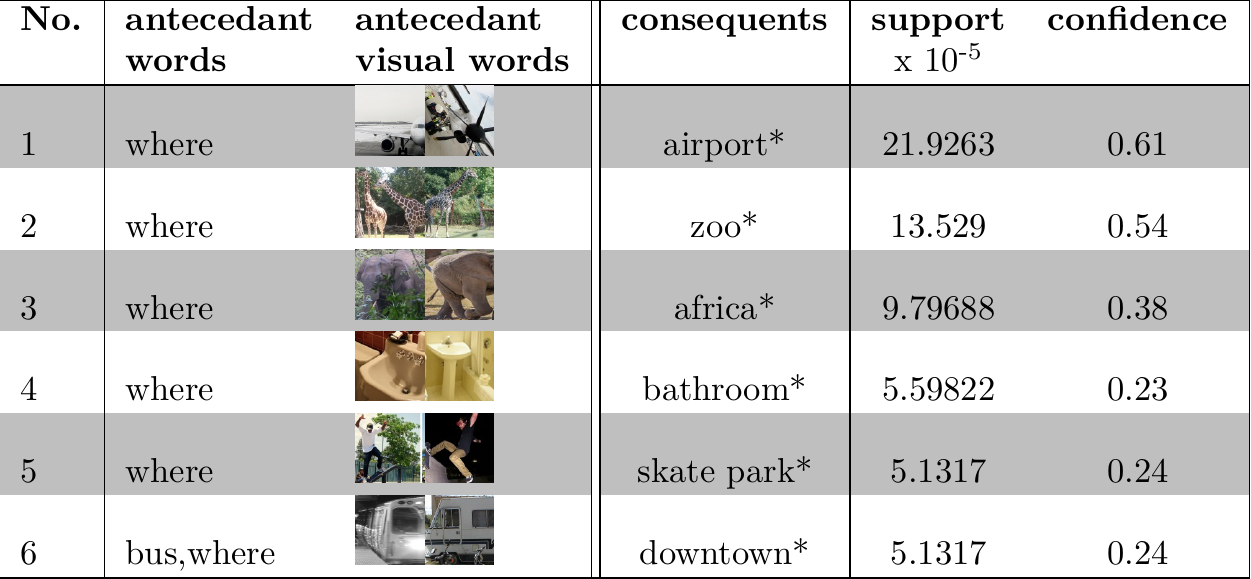}
    \caption{\textbf{Where?} : The model of \cite{KazemiE17} has learned that giraffes can be found in zoos, elephants are from Africa, aircraft can be found in airports and that buses are found in the downtown of a city}
    \label{fig:where}
\end{figure*}

After applying the method of Section \ref{sec:method}, we will examine some rules that have been learned by our method on some popular question types in \vqa. Question types are taken from \cite{antol2015vqa} and for the purpose of brevity, only a very few instructive rules for each question type are displayed. These question types are : ``What is he/she doing?"\ref{sec:doing}, ``Where?" (Figure \ref{fig:where}), ``How many?" (Section \ref{sec:many}), ``What brand?" (Figure \ref{fig:brand}), and ``Why?"(Section \ref{sec:why}). The tables we present are to be interpreted thus : A question containing the antecedent words paired with an image containing the antecedent visual words can sometimes (but not always) lead to the consequent answer. Two instances of patches mapping to this visual word (Section \ref{sec:codebook}) are provided. The presence of an $*$ after the consequent is to remind the reader that the consequent word came from the set of answers. 

\begin{figure*}
    \includegraphics[width=\textwidth,height=.27\textheight]{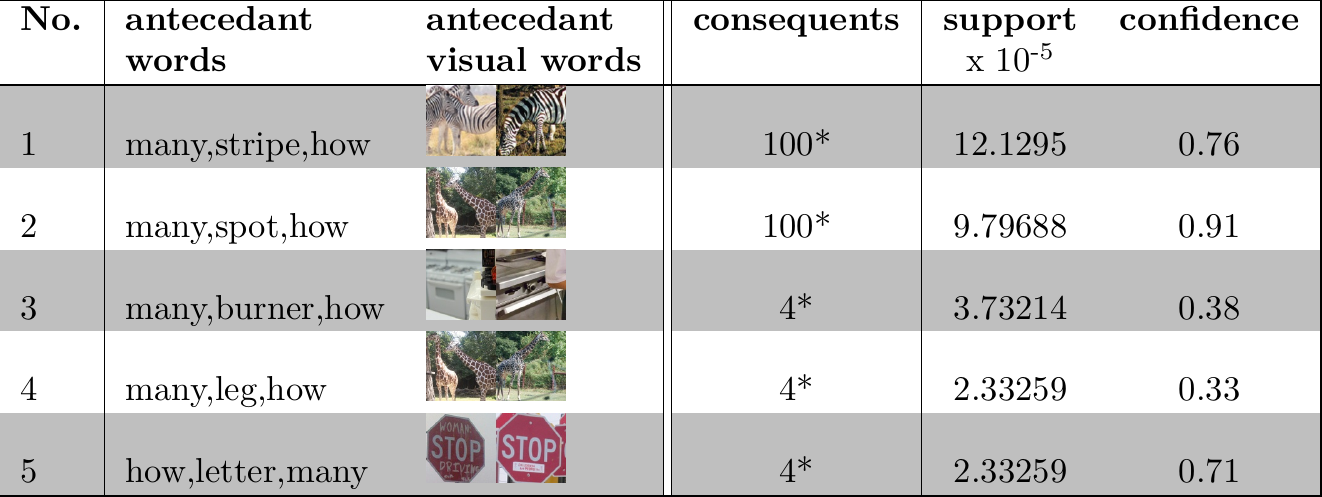}
    \caption{\textbf{How many?} : Rule 3-5 show that stoves have 4 burners, giraffes have 4 legs and stop signs have 4 letters. Giraffes and zebras have many (100) spots and stripes, respectively (rules 1-2).}
    \label{fig:howmany}
\end{figure*}

\subsubsection{How many?}
\label{sec:many}
This particular instance of the trained \vqa model seems to have learned that giraffes have four legs, stop signs have four letters, kitchen stoves have four burners and zebras and giraffes have several (100) stripes and spots respectively (Figure \ref{fig:howmany}). Upon closer examination, we found 33 questions (out of \textgreater 200k) in the \vqa validation set which contain the words \{How,many,burners\} and the most common answer predicted by our model for these is 4 (which also resembles the ground-truth distribution). However, some of them were along the lines of ``How many burners are turned on?", which led to answers different from ``4".

\subsubsection{Why?}
\label{sec:why}
\begin{figure*}
    \includegraphics[width=\textwidth,height=.27\textheight]{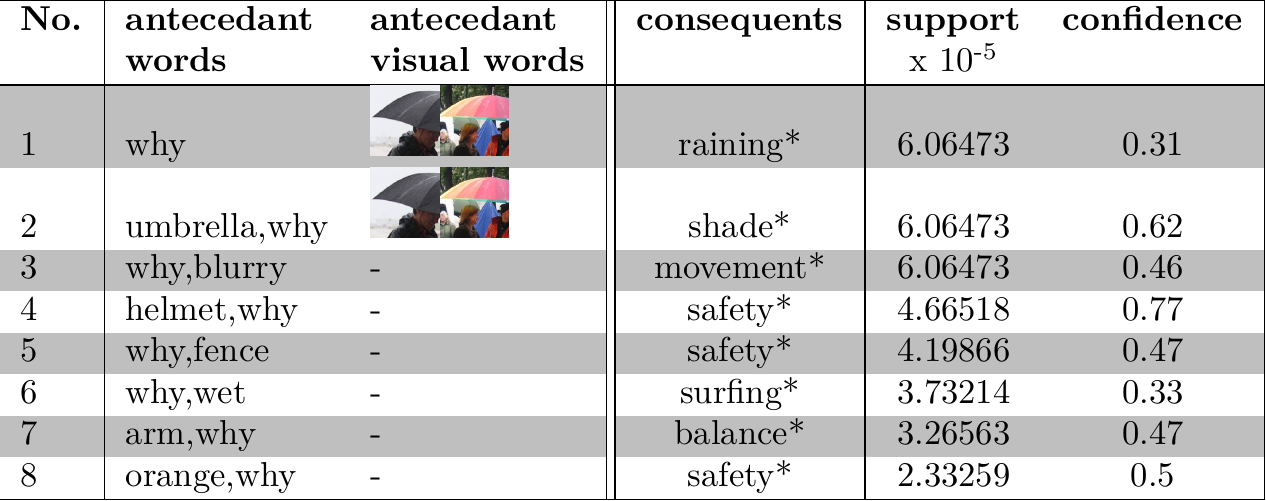}
    \caption{\textbf{Why?} : Rules that exceeded the support threshold indicate that arms are outstretched for balance (rule 7), umbrellas protect one from rain and provide shade (rules 1-2), and that helmets, fences and (wearing) orange lead to safety (rules 4, 5, 8). The absence of visual words in some of these rules indicates that the model is predicting the answer based on question-words only.}
    \label{fig:why}
\end{figure*}

Traditionally, ``Why?" questions in \vqa are considered challenging because they require a reason based answer. We describe some of the rules purportedly learned by our model for answering ``Why?" questions, in Figure \ref{fig:why}. Some interesting but intuitive beliefs that the model has learned are that movements cause blurry photographs ({why,blurry}$\rightarrow$movement), outstretching one's arms help in balancing ({why,arm}$\rightarrow$balance) and that people wear helmets or orange vests for the purpose of safety ({why,helmet/orange}$\rightarrow$safety). In many of these cases, no visual element has been picked up by the rule mining algorithm - this strongly indicates that the models are memorizing the answers to the ``Why?" questions, and not performing any reasoning. In other words, we could ask the question ``Why is the photograph blurry?" to an irrelevant image and obtain ``Movement" as the predicted answer.

\subsubsection{What is he/she doing?}
\label{sec:doing}
\begin{figure*}[!b]
    \includegraphics[width=\textwidth,height=.26\textheight]{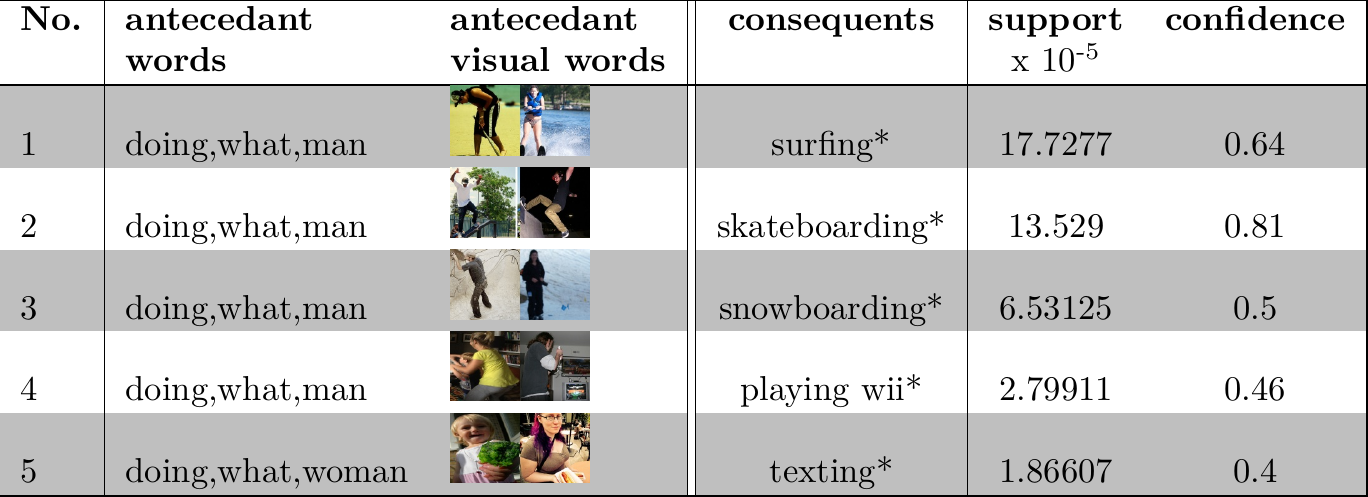}
    \caption{\textbf{What is he/she doing?} : We observed a difference in diversity of rules for male (skateboarding, snowboarding, surfing) and female pronouns (texting) even at very low support. This indicates that the \vqa, or more likely, the MSCOCO datasets are unintentionally skewed in terms of gender.}
    \label{fig:doing}
\end{figure*}

More interesting are our results on the ``What is he/she doing?" category of questions (Figure \ref{fig:doing}). While common activities like ``snowboarding" or ``surfing" are prevalant among the answers, we noticed a difference in rules learned for male and female pronouns. For the female pronoun (she/woman/girl/lady), we observed only stereotypical outputs like ``texting" even for a very low support, as compared to a more diverse set of responses with the male pronoun. This is likely, a reflection on the inherent bias of the MSCOCO dataset which the \vqa dataset of \cite{antol2015vqa, GoyalKSBP17} is based on. Curiously, another work by \cite{HendrixSnowboard} had similar observations for image captioning models also based on MSCOCO.

\begin{figure*}
    \includegraphics[width=\textwidth]{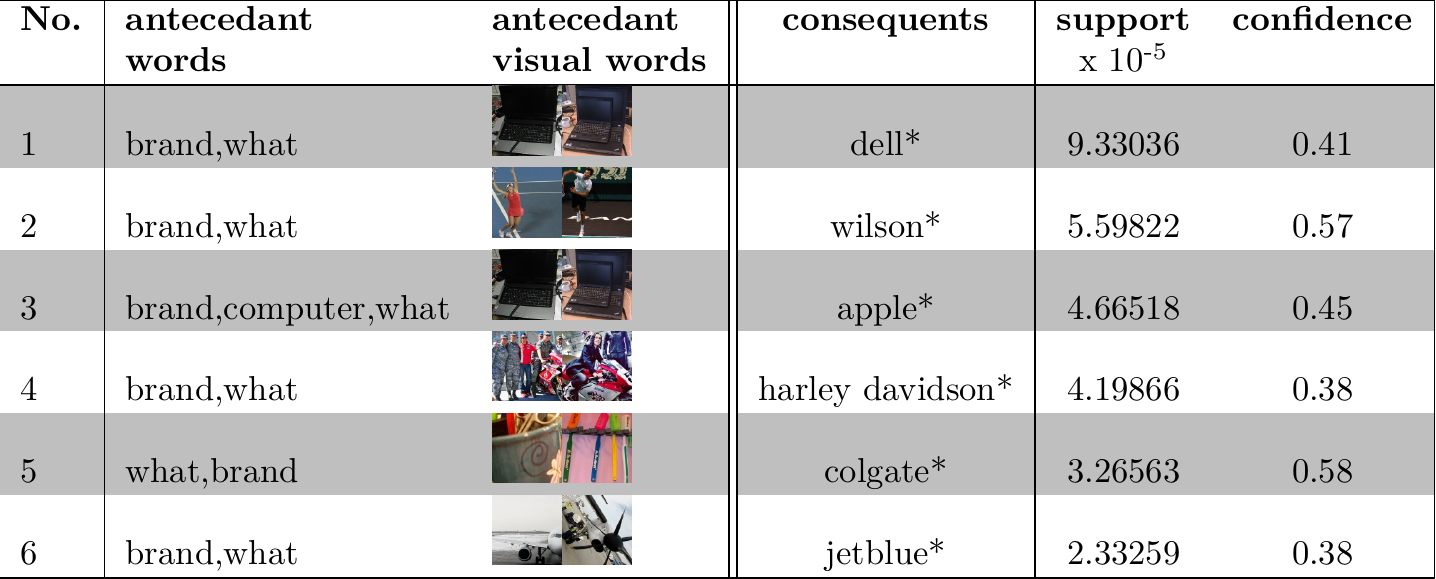}
    \caption{\textbf{What brand?} : The \vqa model seems to have learned that the Wilson brand is related to tennis, Dell and Apple make laptop computers and that Jetblue is a ``brand" of airline. 
    }
    \label{fig:brand}
\end{figure*}

\section{Limitations}
\label{sec:limitations}
While simplicity is the primary advantage of our method, some drawbacks are the following : the exact nature of the rules is limited by the process used to generate the visual vocabulary. In other words, while our method provides a unique insight into the behavior of a \vqa model, there surely exist some rules that the models seem to follow which cannot be captured by this method. For example, rules involving colors are difficult to identify because ResNets are trained to be somewhat invariant to colors, so purely color-based visual words are hard to compute. Other examples include inaccurate visual code-words - for example, in rule 4 of Figure \ref{fig:brand}, the antecedant visual word does show a motorbike, although not a Harley Davidson. Similarly a code-word contains images of scissors and toothbrushes grouped together as part of the (What,brand$\rightarrow$Colgate) associate rule (rule 5 of Figure \ref{fig:brand}). 

\section{Conclusion}
\label{sec:conclusion}
In this work, we present a simple technique to explicitly discover biases and correlations learned by \vqa models. To do so, we store in a database - the words in the question, the response of the model to the question and the portion of the image attended to by the model. Our method then leverages the Apriori algorithm\cite{AgrawalAPriori} to discover rules from this database. We glean from our experiments that \vqa models intuitively seem to correlate \emph{elements} (both textual and visual) in the question and image to answers. 

Our work is consistent with prior art in machine learning on fairness and accountability\cite{HendrixSnowboard}, which often shows a skew towards one set of implied factors (like gender), compared to others. It is also possible to use the ideas in this work to demonstrate effectiveness of VQA systems - showing dataset biases presented by a frequent itemset and rule miner is a middle-ground between quantitative and qualitative results. Finally, our method is not limited only to \vqa, but any problem with a discrete vocabulary. A possible future extension of this work is to track the development of these rules as a function of training time.

{\small
\bibliographystyle{ieee}
\bibliography{egbib}
}

\end{document}